\documentclass[conference]{IEEEtran}
\IEEEoverridecommandlockouts

\usepackage{cite}
\usepackage{amsmath,amssymb,amsfonts}
\usepackage{algorithmic}
\usepackage{graphicx}
\usepackage{textcomp}
\usepackage{xcolor}
\usepackage{multirow}

\def\BibTeX{{\rm B\kern-.05em{\sc i\kern-.025em b}\kern-.08em
    T\kern-.1667em\lower.7ex\hbox{E}\kern-.125emX}}
\begin{document}

\title{Building Robust and Scalable Multilingual ASR for Indian Languages\\}

\author{\IEEEauthorblockN{Arjun Gangwar}
\IEEEauthorblockA{\textit{Wadhwani School of Data Science \& AI} \\
\textit{Indian Institute of Technology, Madras}\\
Chennai, India \\
arjungangwar@gmail.com}
\and
\IEEEauthorblockN{Kaousheik Jayakumar}
\IEEEauthorblockA{\textit{Department of Electrical Engineering} \\
\textit{Indian Institute of Technology, Madras}\\
Chennai, India \\
kaousheik@gmail.com}
\and
\IEEEauthorblockN{S. Umesh}
\IEEEauthorblockA{\textit{Department of Electrical Engineering} \\
\textit{Indian Institute of Technology, Madras}\\
Chennai, India \\
umeshs@ee.iitm.ac.in}
}

\maketitle

\begin{abstract}
This paper describes the systems developed by SPRING Lab, Indian Institute of Technology Madras, for the ASRU MADASR 2.0 challenge. The systems developed focuses on adapting ASR systems to improve in predicting the language and dialect of the utterance among 8 languages across 33 dialects. We participated in Track 1 and Track 2, which restricts the use of additional data and develop from-the-scratch multilingual systems. We presented a novel training approach using Multi-Decoder architecture with phonemic Common Label Set (CLS) as intermediate representation. It improved the performance over the baseline (in the CLS space). We also discuss various methods used to retain the gain obtained in the phonemic space while converting them back to the corresponding grapheme representations. Our systems beat the baseline in 3 languages (Track 2) in terms of WER/CER and achieved the highest language ID and dialect ID accuracy among all participating teams (Track 2).
\end{abstract}


\section{\textbf{Introduction}}

The ASRU MADASR 2.0\footnote{https://sites.google.com/view/respinasrchallenge2025/home} challenge presents a multilingual, multi-dialect dataset spanning over 8 low-resource Indian languages. The challenge provides a dataset of 1200 hours - 150 hours per language - and is evaluated over 4 tracks. The ASR systems developed are evaluated on hidden test sets with metrics such as word error rate (WER), character error rate (CER), language ID accuracy (LID accuracy) and dialect ID accuracy (DID accuracy). These are the 4 tracks with different levels of restrictions.
\begin{itemize}
    \item Track 1 allows only the use of given 30 hours small subset per language with no no external data or models.
    \item Track 2 allows only the use of given 120 hours large subset per language with no external data or models.
    \item Track 3 - an extension of track 1 with the usage of any external data or pre-trained models (open-sourced). 
    \item  Track 4 - an extension of track 2 with the usage of any external data or pre-trained models (open-sourced). 
\end{itemize}
This challenge is among the first works that open sources a labelled language ID and dialect ID tokens inclusive dataset to develop ASR systems for Indian languages. 

This paper presents the ASR systems developed by the SPRING Lab, IIT Madras for tracks 1 and 2. We mainly focus on leveraging the phonemic similarities among Indian languages through a common label set \cite{cls} and discuss ways to retain the gains from the ASR while converting back from the CLS space to corresponding graphemic notations. 


\section{\textbf{Common Label Set}}
\begin{figure}[htbp]
\centerline{\includegraphics[width=0.4\textwidth]{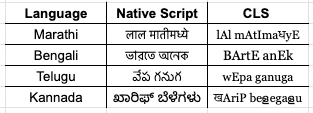}}
\caption{CLS representations of different languages}
\label{cls}
\end{figure}

A very strong correlation exists between phonemes and graphemes of Indian languages, which eases the (grapheme-to-phoneme) G2P conversion. Moreover, the strong phonemic and graphemic similarities between many Indian languages stem from their common roots in related language families. For instance, among the languages used in this challenge, Bhojpuri, Magahi, Marathi, and Chhattisgarhi use the same Devanagari script. Whereas Telugu and Kannada belong to the same Dravidian language family. We exploit these facts by converting the graphemic representations into a common phonemic space using the unified parser \cite{unified_parser}. A standard set of labels are given to phonetically similar speech sounds among different Indian languages known as the Common Label Set. Examples of CLS representation across different languages is shown in Fig~\ref{cls}. Reconstructing text in the native script from CLS representations is inherently challenging due to linguistic phenomena like schwa deletion, geminate correction, and the intricacies of syllable segmentation. This paper discuss a few text-to-text machine transliteration (MT) approaches to retain the gains obtained by the CLS ASR. 

\section{\textbf{Multilingual ASR models}}
We use the ESPnet toolkit \cite{espnet} for all our experiments. Track 1 models are trained on the small dataset (approx. 240 hours), and Track 2 models are trained on the large dataset (approx. 1200 hours). We use 80-dimensional log-Mel Spectrogram as speech features for all our experiments. The Mel Spectrogram are computed using 80 Mel filter banks, with hop length of 10ms and window size of 25ms. All audio files are in 16KHz. We use unified parser to convert native text to CLS. For all our models, language ID and dialect ID information is passed as special token $<LID\_DID>$ in the beginnning of both the CLS and native script text.

\subsection{\textbf{Baseline}}
The baseline system follows a standard encoder-decoder architecture, comprising a Conformer \cite{conformer} encoder and a Transformer decoder \cite{transformer}. It takes log-Mel spectrograms as input to the encoder and generates output in the native script. A special token $<LID>$ is prepended to the target text to indicate language identity. The baseline model does not predict dialects and is trained using a hybrid CTC-Attention loss \cite{hybridctcattn}.

\subsection{\textbf{Approach 1: Cascaded ASR and MT models}}
In this approach, we employ a cascaded system comprising two models. The first model performs speech recognition in the CLS (Common Label Set) space. It takes log-Mel spectrograms as input and generates transcriptions in the CLS format. The second model handles machine transliteration by converting the CLS transcriptions into the target language’s native script. Both the models are trained separately.

For speech recognition, we use an encoder-decoder architecture with a Conformer encoder and a Transformer decoder. The model is trained using a hybrid CTC-Attention loss function. The machine transliteration model also follows an encoder-decoder architecture, utilising transformer layers for both the encoder and decoder, and is trained using only the attention-based cross-entropy loss.

\subsection{\textbf{Approach 2: Multi-Decoder model}}
\begin{figure}[htbp]
\centerline{\includegraphics[width=0.5\textwidth]{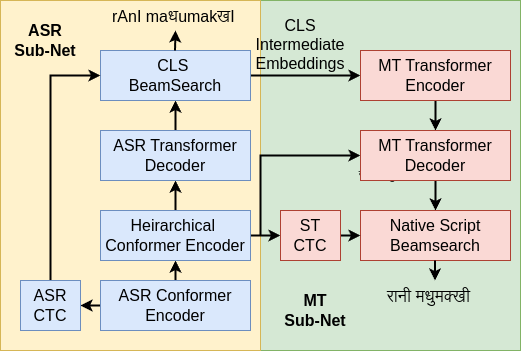}}
\caption{Multi-decoder architecture}
\label{multi-decoder}
\end{figure}

The multi-decoder architecture \cite{multidecoder} consists of two sub-networks as shown in the Fig~\ref{multi-decoder}: an ASR sub-network and an MT (machine transliteration) sub-network. The ASR sub-network employs a Conformer encoder followed by a Transformer decoder, while the MT sub-network uses a lightweight Transformer encoder and a Transformer decoder.

The ASR sub-network takes log-Mel spectrograms as input and generates hidden states through its decoder. These hidden states are passed directly to the encoder of the MT sub-network. The MT decoder then produces the final output in the target language’s native script. The ASR sub-network is trained using a hybrid CTC-Attention loss, where the CTC loss is computed with respect to the CLS (Common Label Set) transcription. The MT sub-network is trained using an attention-based cross-entropy loss. Both the sub-nets are jointly trained.
Additionally, the MT decoder incorporates cross-attention over the speech encoder outputs, allowing it to access acoustic information directly. This auxiliary speech context helps correct errors in the ASR output during transliteration. An optional hierarchical encoder can also be inserted on top of the speech encoder, comprising additional Transformer layers and enabling an auxiliary CTC loss computed with native script transcriptions. This hierarchical setup has been shown to improve transliteration performance.

The multi-decoder architecture retains the benefits of a cascaded system such as the ability to perform beam search and re-scoring on intermediate ASR outputs while reducing the error propagation typically associated with cascaded pipelines.

\subsubsection{\textbf{From Scratch training}}
In this approach, the weights of both sub-networks are randomly initialized. We observed that the ASR sub-network requires significantly more training time compared to the MT sub-network. Since the MT task is relatively easier in this setup, the MT sub-network tends to overfit early during training.

\subsubsection{\textbf{ASR initialization}} To mitigate the overfitting of the MT sub-network, we initialize the ASR sub-network with pretrained weights from the cascaded ASR model. This alignment helps both sub-networks converge at a similar pace, leading to improved overall performance.

\begin{table*}[t]
\caption{Test results for Track 1}
\centering
\resizebox{\textwidth}{!}{%
\begin{tabular}{|lcccccccccccc|}
\hline
\multicolumn{1}{|c|}{\multirow{2}{*}{\textbf{Models/Languages}}} & \multicolumn{8}{c|}{\textbf{Language-Wise CER\%}} & \multicolumn{1}{c|}{\multirow{2}{*}{\textbf{\begin{tabular}[c]{@{}c@{}}Average \\ CER \%\end{tabular}}}} & \multicolumn{1}{c|}{\multirow{2}{*}{\textbf{\begin{tabular}[c]{@{}c@{}}Average \\ WER \%\end{tabular}}}} & \multicolumn{1}{c|}{\multirow{2}{*}{\textbf{LID \%}}} & \multirow{2}{*}{\textbf{DID \%}} \\ \cline{2-9}
\multicolumn{1}{|c|}{} & \multicolumn{1}{l|}{Bhojpuri (bh)} & \multicolumn{1}{l|}{Bengali (bn)} & \multicolumn{1}{l|}{Chhattisgarhi (ch)} & \multicolumn{1}{l|}{Kannada (kn)} & \multicolumn{1}{l|}{Magahi (mg)} & \multicolumn{1}{l|}{Marathi (mr)} & \multicolumn{1}{l|}{Maithili (mt)} & \multicolumn{1}{l|}{Telugu (te)} & \multicolumn{1}{c|}{} & \multicolumn{1}{c|}{} & \multicolumn{1}{c|}{} &  \\ \hline
\multicolumn{13}{|c|}{\textbf{Read Speech}} \\ \hline
\multicolumn{1}{|l|}{Baseline with Dialect (Char)} & \multicolumn{1}{c|}{4.56} & \multicolumn{1}{c|}{4.8} & \multicolumn{1}{c|}{3.52} & \multicolumn{1}{c|}{5.45} & \multicolumn{1}{c|}{6.44} & \multicolumn{1}{c|}{4.05} & \multicolumn{1}{c|}{5.27} & \multicolumn{1}{c|}{4.91} & \multicolumn{1}{c|}{4.86} & \multicolumn{1}{c|}{18.7} & \multicolumn{1}{c|}{97.08} & - \\ \cline{1-1}
\multicolumn{1}{|l|}{Multi-Decoder (Char) - ASR Initialized} & \multicolumn{1}{c|}{4.86} & \multicolumn{1}{c|}{5.92} & \multicolumn{1}{c|}{4.25} & \multicolumn{1}{c|}{6.13} & \multicolumn{1}{c|}{7.01} & \multicolumn{1}{c|}{4.64} & \multicolumn{1}{c|}{6.04} & \multicolumn{1}{c|}{5.76} & \multicolumn{1}{c|}{5.57} & \multicolumn{1}{c|}{21.09} & \multicolumn{1}{c|}{96.80} & 70.80 \\ \hline
\multicolumn{13}{|c|}{\textbf{Spontaneous Speech}} \\ \hline
\multicolumn{1}{|l|}{Baseline with Dialect (Char)} & \multicolumn{1}{c|}{26.08} & \multicolumn{1}{c|}{25.64} & \multicolumn{1}{c|}{20.52} & \multicolumn{1}{c|}{30.74} & \multicolumn{1}{c|}{27.32} & \multicolumn{1}{c|}{16.06} & \multicolumn{1}{c|}{27.33} & \multicolumn{1}{c|}{26.28} & \multicolumn{1}{c|}{25.39} & \multicolumn{1}{c|}{61.7} & \multicolumn{1}{c|}{79.94} & - \\ \cline{1-1}
\multicolumn{1}{|l|}{Multi-Decoder (Char) - ASR Initialized} & \multicolumn{1}{c|}{29.63} & \multicolumn{1}{c|}{39.03} & \multicolumn{1}{c|}{23.29} & \multicolumn{1}{c|}{40.39} & \multicolumn{1}{c|}{30.27} & \multicolumn{1}{c|}{17.82} & \multicolumn{1}{c|}{30.1} & \multicolumn{1}{c|}{29.7} & \multicolumn{1}{c|}{30.63} & \multicolumn{1}{c|}{67.01} & \multicolumn{1}{c|}{72.68} & 29.08 \\ \hline
\end{tabular}%
}
\label{table2}
\end{table*}

\begin{table*}[t]
\caption{Test results for Track 2}
\centering
\resizebox{\textwidth}{!}{%
\begin{tabular}{|lccccccrrcccc|}
\hline
\multicolumn{1}{|c|}{\multirow{2}{*}{\textbf{Models/Languages}}} & \multicolumn{8}{c|}{\textbf{Language-Wise CER\%}} & \multicolumn{1}{c|}{\multirow{2}{*}{\textbf{\begin{tabular}[c]{@{}c@{}}Average \\ CER \%\end{tabular}}}} & \multicolumn{1}{c|}{\multirow{2}{*}{\textbf{\begin{tabular}[c]{@{}c@{}}Average \\ WER \%\end{tabular}}}} & \multicolumn{1}{c|}{\multirow{2}{*}{\textbf{LID \%}}} & \multirow{2}{*}{\textbf{DID \%}} \\ \cline{2-9}
\multicolumn{1}{|c|}{} & \multicolumn{1}{l|}{Bhojpuri (bh)} & \multicolumn{1}{l|}{Bengali (bn)} & \multicolumn{1}{l|}{Chhattisgarhi (ch)} & \multicolumn{1}{l|}{Kannada (kn)} & \multicolumn{1}{l|}{Magahi (mg)} & \multicolumn{1}{l|}{Marathi (mr)} & \multicolumn{1}{l|}{Maithili (mt)} & \multicolumn{1}{l|}{Telugu (te)} & \multicolumn{1}{c|}{} & \multicolumn{1}{c|}{} & \multicolumn{1}{c|}{} &  \\ \hline
\multicolumn{13}{|c|}{\textbf{Read Speech}} \\ \hline
\multicolumn{1}{|l|}{Baseline with Dialect (Char)} & \multicolumn{1}{c|}{4.14} & \multicolumn{1}{c|}{4.19} & \multicolumn{1}{c|}{3.15} & \multicolumn{1}{c|}{4.7} & \multicolumn{1}{c|}{5.63} & \multicolumn{1}{c|}{3.28} & \multicolumn{1}{r|}{5.04} & \multicolumn{1}{r|}{4.4} & \multicolumn{1}{c|}{4.3} & \multicolumn{1}{c|}{16.92} & \multicolumn{1}{c|}{96.03} & - \\ \cline{1-1}
\multicolumn{1}{|l|}{Multi-Decoder (Char) - ASR Initialized} & \multicolumn{1}{c|}{\textbf{4.11}} & \multicolumn{1}{c|}{4.6} & \multicolumn{1}{c|}{3.24} & \multicolumn{1}{c|}{\textbf{4.93}} & \multicolumn{1}{c|}{\textbf{5.57}} & \multicolumn{1}{c|}{3.39} & \multicolumn{1}{r|}{\textbf{4.93}} & \multicolumn{1}{r|}{4.47} & \multicolumn{1}{c|}{4.4} & \multicolumn{1}{c|}{17.06} & \multicolumn{1}{c|}{\textbf{97.39}} & 75.36 \\ \hline
\multicolumn{13}{|c|}{\textbf{Spontaneous Speech}} \\ \hline
\multicolumn{1}{|l|}{Baseline with Dialect (Char)} & \multicolumn{1}{c|}{25.5} & \multicolumn{1}{c|}{30.06} & \multicolumn{1}{c|}{19.97} & \multicolumn{1}{c|}{30.86} & \multicolumn{1}{c|}{26.1} & \multicolumn{1}{c|}{14.37} & \multicolumn{1}{r|}{25.59} & \multicolumn{1}{r|}{25.34} & \multicolumn{1}{c|}{25.11} & \multicolumn{1}{c|}{59.01} & \multicolumn{1}{c|}{76.89} & - \\ \cline{1-1}
\multicolumn{1}{|l|}{Multi-Decoder (Char) - ASR Initialized} & \multicolumn{1}{c|}{27.08} & \multicolumn{1}{c|}{31.39} & \multicolumn{1}{c|}{20.98} & \multicolumn{1}{c|}{33.1} & \multicolumn{1}{c|}{27.55} & \multicolumn{1}{c|}{15.26} & \multicolumn{1}{r|}{28.68} & \multicolumn{1}{r|}{28.63} & \multicolumn{1}{c|}{26.99} & \multicolumn{1}{c|}{62.32} & \multicolumn{1}{c|}{\textbf{77.61}} & 33.33 \\ \hline
\end{tabular}%
}
\label{table3}
\end{table*}

\begin{table*}[t]
\caption{Dev set results for Track 1}
\centering
\resizebox{\textwidth}{!}{%
\begin{tabular}{|l|rrrrrrrr|r|r|r|r|}
\hline
\multicolumn{1}{|c|}{\multirow{2}{*}{\textbf{Models/Languages}}} & \multicolumn{8}{c|}{\textbf{Language-Wise WER\%}} & \multicolumn{1}{c|}{\multirow{2}{*}{\textbf{\begin{tabular}[c]{@{}c@{}}Average \\ CER \%\end{tabular}}}} & \multicolumn{1}{c|}{\multirow{2}{*}{\textbf{\begin{tabular}[c]{@{}c@{}}Average \\ WER \%\end{tabular}}}} & \multicolumn{1}{c|}{\multirow{2}{*}{\textbf{LID \%}}} & \multirow{2}{*}{\textbf{DID \%}} \\ \cline{2-9}
\multicolumn{1}{|c|}{} & \multicolumn{1}{l|}{Bhojpuri (bh)} & \multicolumn{1}{l|}{Bengali (bn)} & \multicolumn{1}{l|}{Chhattisgarhi (ch)} & \multicolumn{1}{l|}{Kannada (kn)} & \multicolumn{1}{l|}{Magahi (mg)} & \multicolumn{1}{l|}{Marathi (mr)} & \multicolumn{1}{l|}{Maithili (mt)} & \multicolumn{1}{l|}{Telugu (te)} & \multicolumn{1}{c|}{} & \multicolumn{1}{c|}{} & \multicolumn{1}{c|}{} & \multicolumn{1}{c|}{} \\ \hline
Baseline with Dialect (Char) & \multicolumn{1}{r|}{15.12} & \multicolumn{1}{r|}{19.73} & \multicolumn{1}{r|}{12.39} & \multicolumn{1}{r|}{23.84} & \multicolumn{1}{r|}{19.78} & \multicolumn{1}{r|}{16.32} & \multicolumn{1}{r|}{18.11} & 22.04 & 4.22 & 17.93 & 97.34 & 69.68 \\ \cline{1-1}
CLS ASR (Char) & \multicolumn{1}{r|}{15.25} & \multicolumn{1}{r|}{19.41} & \multicolumn{1}{r|}{12.54} & \multicolumn{1}{r|}{23.52} & \multicolumn{1}{r|}{19.91} & \multicolumn{1}{r|}{16.17} & \multicolumn{1}{r|}{17.17} & 22.38 & 3.92 & 17.78 & \multirow{2}{*}{97.45} & \multirow{2}{*}{70.3} \\
+ MT (BPE) & \multicolumn{1}{r|}{17.29} & \multicolumn{1}{r|}{28.88} & \multicolumn{1}{r|}{14.21} & \multicolumn{1}{r|}{23.8} & \multicolumn{1}{r|}{21.4} & \multicolumn{1}{r|}{16.26} & \multicolumn{1}{r|}{18.29} & 22.64 & 5.22 & 19.99 &  &  \\ \cline{1-1}
Multi-Decoder (BPE) - From Scratch & \multicolumn{1}{r|}{16.61} & \multicolumn{1}{r|}{27.85} & \multicolumn{1}{r|}{13.4} & \multicolumn{1}{r|}{23.48} & \multicolumn{1}{r|}{20.76} & \multicolumn{1}{r|}{15.68} & \multicolumn{1}{r|}{18.45} & 22.71 & 5.02 & 19.86 & 97.39 & 70.55 \\ \cline{1-1}
Multi-Decoder (Char) - From Scratch & \multicolumn{1}{r|}{16.44} & \multicolumn{1}{r|}{22.09} & \multicolumn{1}{r|}{12.52} & \multicolumn{1}{r|}{25.75} & \multicolumn{1}{r|}{21.86} & \multicolumn{1}{r|}{19.43} & \multicolumn{1}{r|}{19.4} & 24.79 & 4.72 & 19.69 & 96.72 & 65.73 \\ \cline{1-1}
Multi-Decoder (Char) - ASR Initialized & \multicolumn{1}{r|}{15.88} & \multicolumn{1}{r|}{22.18} & \multicolumn{1}{r|}{12.23} & \multicolumn{1}{r|}{25.63} & \multicolumn{1}{r|}{21.25} & \multicolumn{1}{r|}{18.87} & \multicolumn{1}{r|}{19.48} & \multicolumn{1}{r|}{24.6} & \multicolumn{1}{r|}{4.66} & \multicolumn{1}{r|}{19.54} & \multicolumn{1}{r|}{97.21} & \multicolumn{1}{r|}{69.8} \\ \hline
\end{tabular}%
}
\label{table4}
\end{table*}

\begin{table*}[t]
\caption{Dev set results for Track 2}
\centering
\resizebox{\textwidth}{!}{%
\begin{tabular}{|l|rrrrrrrr|r|r|r|r|}
\hline
\multicolumn{1}{|c|}{\multirow{2}{*}{\textbf{Models/Languages}}} & \multicolumn{8}{c|}{\textbf{Language-Wise WER\%}} & \multicolumn{1}{c|}{\multirow{2}{*}{\textbf{\begin{tabular}[c]{@{}c@{}}Average \\ CER \%\end{tabular}}}} & \multicolumn{1}{c|}{\multirow{2}{*}{\textbf{\begin{tabular}[c]{@{}c@{}}Average \\ WER \%\end{tabular}}}} & \multicolumn{1}{c|}{\multirow{2}{*}{\textbf{LID \%}}} & \multirow{2}{*}{\textbf{DID \%}} \\ \cline{2-9}
\multicolumn{1}{|c|}{} & \multicolumn{1}{l|}{Bhojpuri (bh)} & \multicolumn{1}{l|}{Bengali (bn)} & \multicolumn{1}{l|}{Chhattisgarhi (ch)} & \multicolumn{1}{l|}{Kannada (kn)} & \multicolumn{1}{l|}{Magahi (mg)} & \multicolumn{1}{l|}{Marathi (mr)} & \multicolumn{1}{l|}{Maithili (mt)} & \multicolumn{1}{l|}{Telugu (te)} & \multicolumn{1}{c|}{} & \multicolumn{1}{c|}{} & \multicolumn{1}{c|}{} & \multicolumn{1}{c|}{} \\ \hline
Baseline with Dialect (Char) & \multicolumn{1}{r|}{13.71} & \multicolumn{1}{r|}{16.51} & \multicolumn{1}{r|}{9.74} & \multicolumn{1}{r|}{20.57} & \multicolumn{1}{r|}{17.42} & \multicolumn{1}{r|}{13.12} & \multicolumn{1}{r|}{15.67} & 20.73 & 3.52 & 15.31 & 98.37 & 74.87 \\ \cline{1-1}
CLS ASR (Char) & \multicolumn{1}{r|}{13.19} & \multicolumn{1}{r|}{16.05} & \multicolumn{1}{r|}{9.80} & \multicolumn{1}{r|}{20.73} & \multicolumn{1}{r|}{17.04} & \multicolumn{1}{r|}{13.02} & \multicolumn{1}{r|}{14.84} & 18.95 & 3.18 & 14.88 & \multirow{2}{*}{98.13} & \multirow{2}{*}{76.28} \\
+ MT (BPE) & \multicolumn{1}{r|}{15.17} & \multicolumn{1}{r|}{26.07} & \multicolumn{1}{r|}{11.56} & \multicolumn{1}{r|}{20.83} & \multicolumn{1}{r|}{18.59} & \multicolumn{1}{r|}{13.22} & \multicolumn{1}{r|}{16.04} & 19.32 & 4.4 & 17.21 &  &  \\ \cline{1-1}
Multi-Decoder (BPE) - From Scratch & \multicolumn{1}{r|}{15.67} & \multicolumn{1}{r|}{26.67} & \multicolumn{1}{r|}{12.34} & \multicolumn{1}{r|}{22.97} & \multicolumn{1}{r|}{19.27} & \multicolumn{1}{r|}{15.15} & \multicolumn{1}{r|}{18.20} & 22.32 & 4.8 & 18.69 & 97.35 & 72.78 \\ \cline{1-1}
Multi-Decoder (Char) - From Scratch & \multicolumn{1}{r|}{14.53} & \multicolumn{1}{r|}{18.15} & \multicolumn{1}{r|}{10.95} & \multicolumn{1}{r|}{22.57} & \multicolumn{1}{r|}{17.87} & \multicolumn{1}{r|}{15.22} & \multicolumn{1}{r|}{15.94} & 21.42 & 3.83 & 16.51 & 97.26 & 71.46 \\ \cline{1-1}
Multi-Decoder (Char) - ASR Initialized & \multicolumn{1}{r|}{13.83} & \multicolumn{1}{r|}{18.24} & \multicolumn{1}{r|}{10.66} & \multicolumn{1}{r|}{22.58} & \multicolumn{1}{r|}{17.63} & \multicolumn{1}{r|}{14.84} & \multicolumn{1}{r|}{16.17} & 21.53 & 3.84 & 16.30 & 97.63 & 74 \\ \hline
\end{tabular}%
}
\label{table5}
\end{table*}

\subsection{\textbf{Implementation Details}}
The ASR encoder in the baseline model follows a Conformer architecture with 8 encoder blocks, each comprising a model dimension of 256, a 1024-dimensional position-wise feed-forward layer, and 4 attention heads. A kernel size of 31 is used, along with the Swish activation function. The ASR decoder is implemented using a Transformer architecture with 6 decoder blocks, a model dimension of 256, a 2048-dimensional feed-forward layer, and 4 attention heads, employing ReLU as the activation function. We use the same configuration for the ASR model in both our cascaded pipeline and the ASR sub-network in the multi-decoder setup. Character-level tokenization is used for the ASR decoder in all cases, whether generating native script in the baseline or CLS outputs in the cascaded pipeline.

The MT model in the cascaded pipeline consists of a Transformer encoder with 6 blocks, each having a model dimension of 512, a 1024-dimensional feed-forward layer, and 4 attention heads. The corresponding Transformer decoder has 6 blocks with a model dimension of 256, a 1024-dimensional feed-forward layer, and 4 attention heads. We employ joint tokenization of both source (CLS) and target (native script) texts using Byte Pair Encoding (BPE) with the SentencePiece library. The vocabulary size is set to 1500 for the small dataset and 3000 for the large dataset.

In the multi-decoder setup, the optional hierarchical encoder shares the same configuration as the ASR encoder described above but uses 6 layers. The MT sub-network consists of a Transformer encoder and decoder, each with 2 blocks, a model dimension of 256, a 2048-dimensional feed-forward layer, and 4 attention heads. We experiment with both character-level and BPE tokenizations in this setup. For the small dataset, we use 500 tokens for CLS and 1500 tokens for the native script. For the large dataset, we use 1000 tokens for CLS and 3000 tokens for the native script.
\vspace{-0.3em}
\section{\textbf{Results and Discussion}}
We start by comparing the results of the baseline with the cascaded CLS ASR + MT pipeline. We see that CLS ASR model comfortably beats the conformer baseline (in CLS space). This is because of the fewer number of target outputs for the decoder to predict, resulting in less confusion. But while converting back to the native-script space we see a drop in performance due to errors propagating from ASR to MT.

To reduced the errors being cascaded, we implement a multi-decoder approach trained on BPE and char tokens. The char model outperforms the BPE model due to the nature of the dataset, where the same utterance is spoken in multiple dialects, potentially causing the BPE-based models to overfit. The char-based multi-decoder model performs better than the BPE models across all languages. 

The multi-decoder approach improves over the cascaded pipeline as the loss is not cascaded. But we observe that the MT-subnetwork is overfitted due to the mismatch in saturation times. A multi-decoder network with ASR initialisation allows both sub-networks to saturate at the same time. This is seen in the results tabulated, as the multi-decoder with ASR initialisation outperforms all our models. 

We also see that in Table~\ref{table3},  the LID accuracy improves drastically over the baseline using the multi-decoder approach in both read speech and spontaneous speech subtasks. By splitting tasks, specialising decoders, and using joint end-to-end training, the multi-decoder model avoids conflating transcription in both the CLS and native script domains. This modular approach improves language identification by allowing each component to learn its role more effectively, reducing mislabelled representations due to feedback from multiple decoders, thus enabling clearer cross-stage signals.


\bibliographystyle{IEEEtran}
\bibliography{references.bib}

\vspace{12pt}

\end{document}